\documentclass[10pt,twocolumn,letterpaper]{article}

\usepackage{cvpr}              

\usepackage{amsmath,amssymb}
\usepackage{textcomp} 

\usepackage{booktabs}
\usepackage{multirow}

\usepackage{graphicx}
\usepackage{subcaption}
\usepackage{tikz}
\usetikzlibrary{calc,arrows.meta,positioning}

\usepackage{algorithm}
\usepackage{algorithmic}

\usepackage{xspace}
\newcommand{\lollypop}{\emph{Lollypop}\xspace}

\newcommand{\vect}[1]{\mathbf{#1}}
\newcommand{\norm}[1]{\left\|#1\right\|}

\definecolor{cvprblue}{rgb}{0.21,0.49,0.74}
\usepackage[breaklinks,colorlinks,allcolors=cvprblue]{hyperref}

\title{Robust Camera-to-Mocap Calibration and Verification\\for Large-Scale Multi-Camera Data Capture}

\author{Tianyi Liu \quad Christopher Twigg \quad Patrick Grady \quad Kevin Harris \quad Shangchen Han \quad Kun He \\
Meta\\
{\tt\small \{tianyil, cdtwigg, pgrady, harriskevin, shchhan, kunhe\}@meta.com}
}

\begin{document}
\maketitle
\begin{abstract}
Optical motion capture (mocap) systems are widely used for ground-truth capture in AR/VR, SLAM and robotics datasets. These datasets require extrinsic calibration to align mocap coordinates to external camera frames---a step that is subject to multiple sources of error in practice, and failures often go undetected until they corrupt downstream data. These issues are compounded for fisheye cameras, where spatially non-uniform distortion makes both calibration and verification more challenging. We present a calibration and verification system designed for this setting. Concretely, we target robustness to board-to-marker attachment variation, optimization initialization ambiguity, and session-to-session calibration drift after deployment. The calibration jointly estimates camera extrinsics and the board-to-marker transform, and uses a staged solver to improve convergence reliability under ambiguous initialization. The verification component, \lollypop, provides fast, operator-independent assessment through a measurement chain entirely independent of the calibration data. In experiments on a Meta Quest~3 headset with fisheye cameras, our calibration matches or outperforms Sturm~\etal~\cite{sturm2012benchmark}, and \lollypop reliably detects calibration degradation over time. The system has been deployed and verified in production data collection pipelines.
\end{abstract}

\begin{figure}[t]
  \centering
  \includegraphics[width=0.95\linewidth]{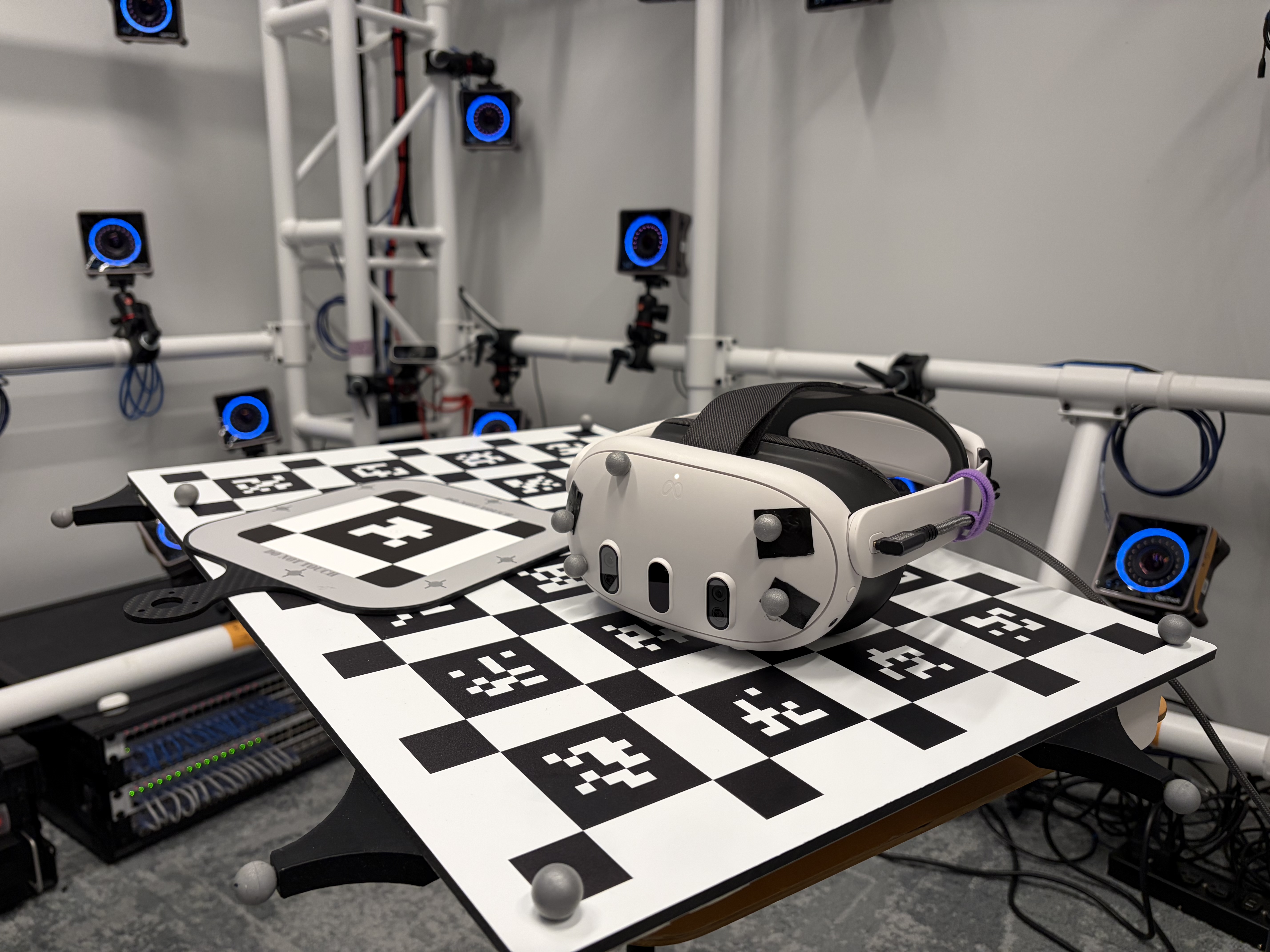}
  \caption{\textbf{System overview.} The calibration setup: a Meta Quest 3 headset (center) and a large ArUco board with retroreflective markers affixed at its corners, placed within a motion capture volume. The board is simultaneously detected in the headset's fisheye camera images and tracked in 3D by the mocap system. Our pipeline jointly estimates camera-to-mocap extrinsics and the board-to-marker transform from this data, then independently verifies calibration quality with Lollypop.}
  \label{fig:system}
\end{figure}

\begin{figure}[t]
  \centering
  \begin{subfigure}{0.48\linewidth}
    \includegraphics[width=\linewidth]{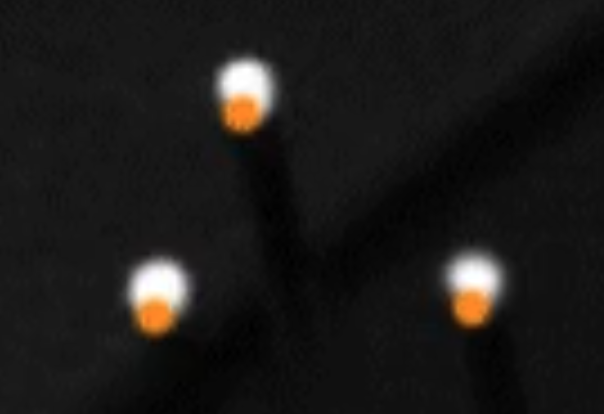}
    \caption{Low-quality calibration}
    \label{fig:badcal}
  \end{subfigure}
  \hfill
  \begin{subfigure}{0.48\linewidth}
    \includegraphics[width=\linewidth,trim=0 8.5bp 0 8.5bp,clip]{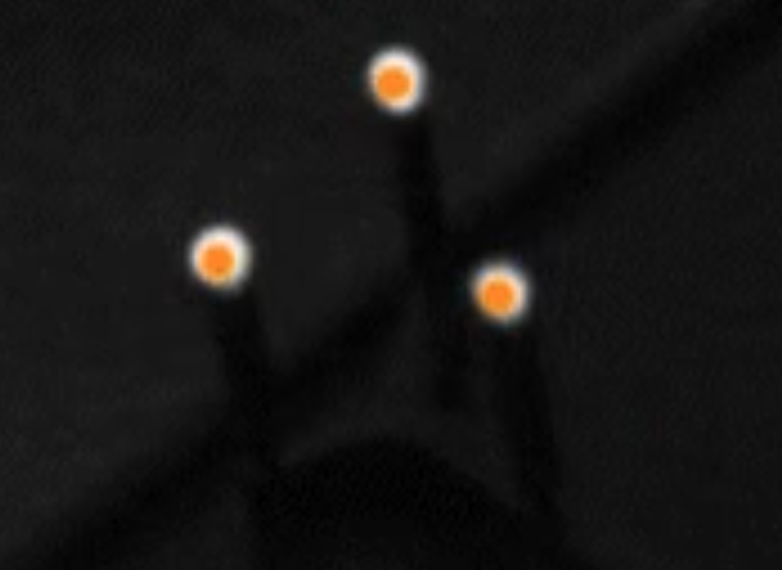}
    \caption{High-quality calibration}
    \label{fig:goodcal}
  \end{subfigure}
  \\[4pt]
  \begin{subfigure}{0.48\linewidth}
    \includegraphics[width=\linewidth,trim=0 5.25bp 0 5.25bp,clip]{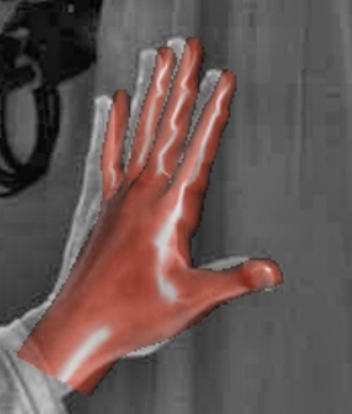}
    \caption{Low-quality calibration}
    \label{fig:hand_bad}
  \end{subfigure}
  \hfill
  \begin{subfigure}{0.48\linewidth}
    \includegraphics[width=\linewidth]{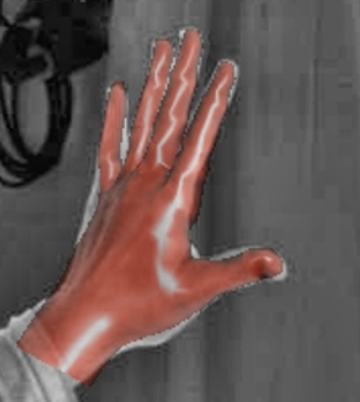}
    \caption{High-quality calibration}
    \label{fig:hand_good}
  \end{subfigure}
  \caption{\textbf{Effect of calibration quality.} Top row: mocap marker reprojections (orange circles) into camera images. Bottom row: reprojected hand-tracking ground truth. A low-quality calibration~(a,\,c) produces visible misalignment; a high-quality calibration~(b,\,d) shows accurate overlay.}
  \label{fig:goodbad}
\end{figure}

\section{Introduction}
\label{sec:intro}

Optical motion capture (mocap) is a standard source of ground-truth pose data for hand tracking~\cite{zimmermann2019freihand,moon2020interhand,fan2023arctic}, visual-inertial SLAM~\cite{schubert2018tumvi,burri2016euroc}, and AR/VR datasets. To use this ground truth in camera images, one must estimate an extrinsic calibration that maps the mocap coordinate frame to each camera frame. This calibration step is often under-documented in prior work~\cite{zimmermann2019freihand,moon2020interhand,fan2023arctic,schubert2018tumvi,burri2016euroc}: calibration quality is typically assumed rather than explicitly validated. In practice, calibration is vulnerable to several sources of error. Capture systems are operated by humans, hardware is handled between sessions, and the physical geometry can change without being reflected in standard quality metrics. These errors can silently propagate into corrupted ground truth, degraded models, and irreproducible results.

The calibration itself is a hand-eye problem~\cite{tsai1989new}: given a camera and a mocap system that both observe a moving calibration board with attached retroreflective markers, estimate the rigid transform from the mocap world frame to each camera frame.
A central challenge is the \emph{board-to-marker transform}, i.e., the rigid offset between the calibration pattern coordinate frame and the attached mocap rigid body.
Standard approaches assume this transform is precisely known~\cite{sturm2012benchmark}. In practice, this is a fragile assumption: retroreflective markers are reattached, shifted, or replaced between sessions, and the physical relationship between markers and the calibration pattern can drift over time without detection.
For fisheye cameras, common in AR/VR headsets and SLAM systems, strong spatially non-uniform distortion further complicates both calibration and verification. \Cref{fig:goodbad} shows examples where inaccurate calibration produces visibly misaligned reprojected mocap markers and hand-tracking ground truth.

Verification is equally problematic. The standard quality metric---board reprojection error---is the residual of the calibration optimization itself, computed on the same data used to estimate the parameters. It is inherently self-referential: a calibration can report low board RMS while harboring significant errors in regions not probed by the calibration data. The alternative is visual inspection (\cref{fig:goodbad}), where an operator examines reprojected markers and subjectively judges alignment. This is subjective, inconsistent across operators, and does not scale.

We present a robust calibration and verification system designed for large-scale data collection. In this paper, robustness has three concrete meanings: (i) robustness to board-to-marker attachment variation across sessions, (ii) robustness of optimization to ambiguous or poor initialization, and (iii) robustness of calibration quality control to deployment-time drift. The calibration jointly estimates camera extrinsics and the board-to-marker transform, and uses a staged optimization pipeline to improve convergence reliability under ambiguous initialization. The verification component, \lollypop, uses a separate recording with a fiducial-marker/mocap rigid body to test the calibration through a measurement chain entirely independent of the calibration data, providing a fast, objective, operator-independent pass/fail decision. The system has been deployed in production capture pipelines.

\noindent Our contributions are:
\begin{enumerate}
    \item A joint calibration that removes the known-board-to-marker assumption, making calibration robust to marker placement variation across sessions. Benchmarked against Sturm~\etal~\cite{sturm2012benchmark}.
    \item A staged solver (random-restart Procrustes, 3D refinement, and 2D reprojection refinement) that improves convergence reliability under difficult initialization.
    \item \lollypop, a verification device and metric providing fast, operator-independent calibration assessment through an independent measurement chain.
\end{enumerate}

\begin{figure}[!tbp]
  \centering
  \includegraphics[width=\linewidth]{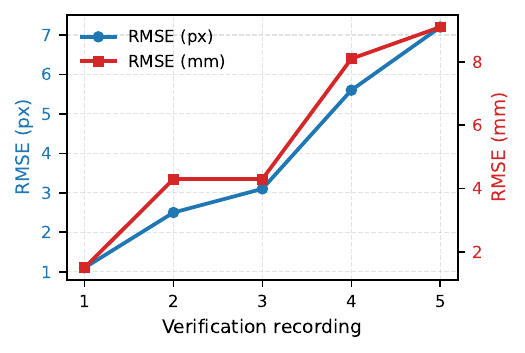}
  \caption{\textbf{Calibration drift over repeated use.} After an initial calibration, the headset is repeatedly donned and doffed, and the mocap markers are lightly perturbed to emulate real-world capture usage. Five intermediate verification recordings are evaluated with \lollypop. The pixel-domain and metric-domain RMSE both increase, indicating progressive calibration degradation.}
  \label{fig:drift}
\end{figure}
\section{Related Work}
\label{sec:related}

\paragraph{Camera calibration and fisheye models.}
The foundational work for camera calibration using planar targets was established by Zhang~\cite{zhang2000flexible}, building on earlier work by Tsai~\cite{tsai1987versatile} that first characterized radial distortion. For fisheye and wide-angle lenses, specialized distortion models have been developed. A widely adopted approach is the generic polynomial model of Kannala and Brandt~\cite{kannala2006generic}, which forms the basis of the Fisheye62 model used in this work. Other popular methods, such as the OCamCalib toolbox by Scaramuzza~\etal~\cite{scaramuzza2006toolbox}, model the projection function as a Taylor series expansion, providing a flexible framework for a wide range of omnidirectional cameras. More recently, Tezaur~\etal~\cite{tezaur2022new} proposed a non-central model that extends Scaramuzza's work to account for angle-dependent viewpoint shifts, addressing a key limitation of central models where the single viewpoint assumption can be violated in fisheye cameras, especially when observing objects at close range.

\paragraph{Camera-to-mocap calibration.}
Calibrating a camera to an external tracking system is a form of hand-eye calibration. Tsai and Lenz~\cite{tsai1989new} formulated the single-unknown case as $AX = XB$; Zhuang~\etal~\cite{zhuang1994simultaneous} extended this to the two-unknown form $AX = YB$, jointly solving for the robot-world and tool-flange transforms. Dornaika and Horaud~\cite{dornaika1998simultaneous} proposed a nonlinear optimization formulation for the same problem. In the camera-mocap setting, $X$ corresponds to the board-to-marker transform and $Y$ to the camera extrinsic. Sturm~\etal~\cite{sturm2012benchmark} simplify the problem by assuming $X$ is precisely known, reducing to a single unknown per camera. Chiodini~\etal~\cite{chiodini2018camera} address the camera-to-mocap problem directly by using the mocap system's own retroreflective spherical markers as the calibration target, solving for the camera-to-mocap transform via P3P-RANSAC followed by nonlinear reprojection-error minimization, and providing a Monte Carlo uncertainty analysis of the result. However, like Sturm~\etal, they provide no independent means of verifying calibration quality after the fact.

The known-board-to-marker assumption is practical when markers can be precisely placed, but becomes fragile in production environments where markers are reattached between sessions, boards are rebuilt, or the board-to-marker relationship drifts over time. Our approach treats $X$ as a free variable and solves for it jointly with $Y$, removing this assumption.

\begin{figure}[t]
  \centering
  \includegraphics[width=0.95\linewidth]{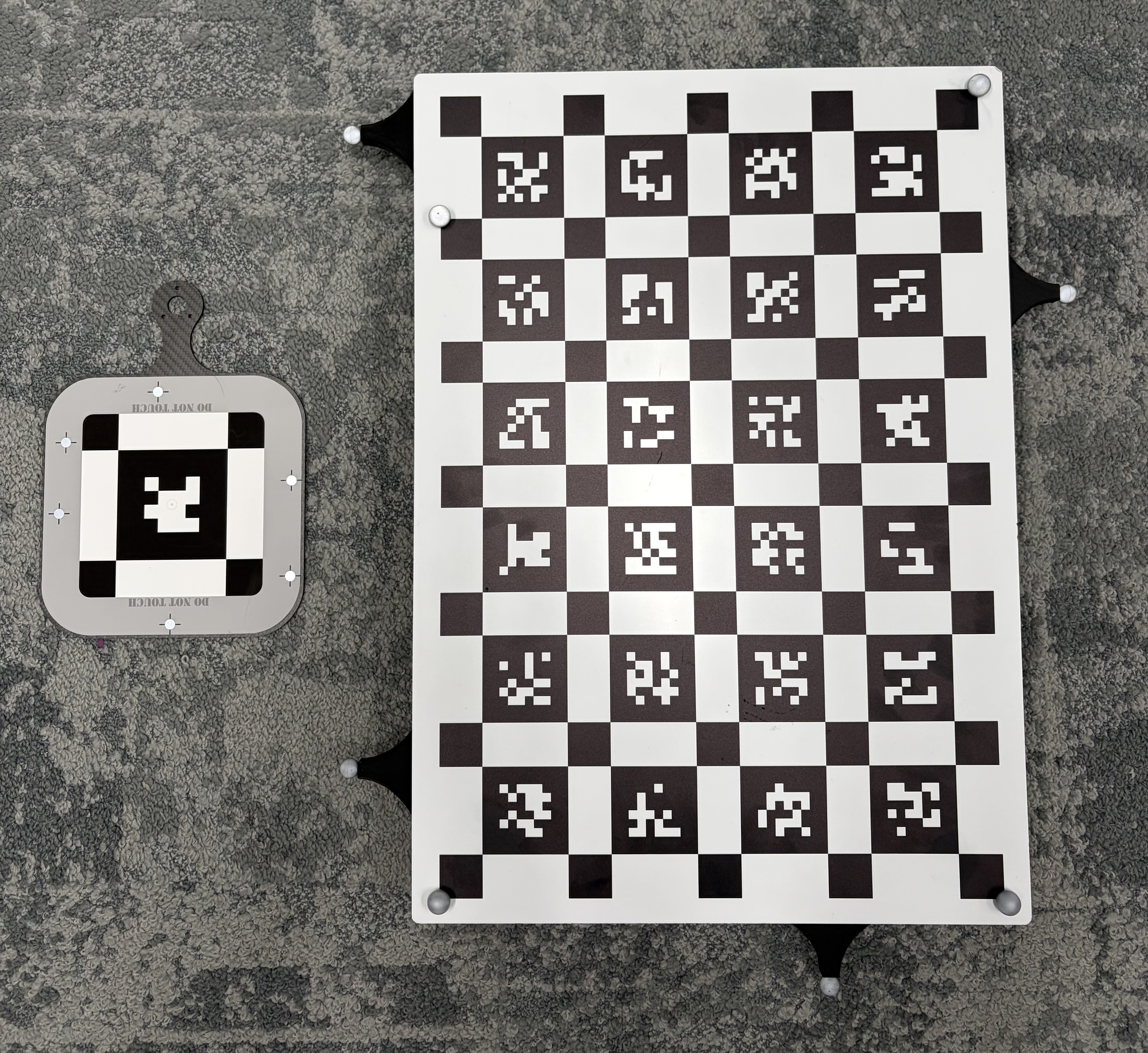}
  \caption{\textbf{Calibration board and \lollypop device.} Left: the \lollypop verification device (single ArUco marker with retroreflective mocap markers). Right: the custom ArUco calibration board with mocap markers attached at the corners and edges. The board-to-marker transform $X$ relates the printed pattern frame to the mocap rigid body frame; our method estimates $X$ jointly rather than assuming it is known.}
  \label{fig:calboard}
\end{figure}

\paragraph{Datasets using camera-mocap calibration.}
Camera-to-mocap calibration underpins ground-truth generation across several application areas. The TUM RGB-D benchmark~\cite{sturm2012benchmark} used a MotionAnalysis system to provide ground-truth poses for an RGB-D sensor; the authors reported extrinsic calibration errors of 3.25--4.03\,mm, but did not discuss the calibration algorithm in detail. In visual-inertial SLAM, TUM-VI~\cite{schubert2018tumvi} and EuRoC~\cite{burri2016euroc} use mocap-derived trajectories as ground truth, with the camera-to-mocap extrinsic determining how trajectory accuracy translates to image-space evaluation.

For human understanding, Human3.6M~\cite{ionescu2014human36m} calibrated four video cameras against a Vicon system, reporting a mean reprojection error of 0.17\,px but their process involved error-prone manual labeling of markers in image frames. Assembly101~\cite{sener2022assembly101} tracks egocentric head pose via a mocap rigid body on a headset; the paper claims ``sub-pixel accuracies'' for the camera rig calibration but provides no specific numerical metrics or repeatability analysis. Hand-object interaction datasets including ARCTIC~\cite{fan2023arctic} use dense Vicon systems (54~cameras) and calibrate using the reflective markers themselves as targets. Across all these works, calibration procedures are described briefly and sometimes rely on manual, error-prone procedures; independent verification of calibration quality---particularly across sessions or in the presence of real-world perturbations---is also not addressed.

\paragraph{Calibration quality verification.}
The most common quality metric is reprojection error on the calibration data itself~\cite{zhang2000flexible}---the residual of the optimization and therefore inherently self-referential. This limitation extends to camera-to-mocap calibration; for example, Schofield~\etal~\cite{schofield2018calibration} estimate camera-mocap extrinsics using retroreflective markers but verify the result solely via the reprojection error of those same markers. Hold-out validation provides partial independence but uses the same target and detection pipeline. Richardson~\etal~\cite{richardson2013aprilcal} present AprilCal to improve repeatability, but do not address independent post-calibration verification against an external reference. Multi-view consistency checks~\cite{furgale2013unified} verify internal consistency but cannot detect systematic errors shared across views. As Shu~\etal~\cite{shu2024spatiotemporal} note, evaluating hand-eye (camera-to-mocap) calibration in real-world systems is inherently difficult because the true physical offset is rarely available, forcing reliance on indirect proxy metrics like trajectory alignment. These approaches share a common limitation: they evaluate calibration quality using the same measurement chain---target, detector, data distribution---employed during estimation. Few provide a verification protocol that is fully independent of the calibration procedure.

Our work addresses both axes: a joint estimation formulation robust to board-to-marker uncertainty, paired with an independent verification device and protocol that uses a separate recording and detection pipeline to test the calibration result.

\section{Calibration Method}
\label{sec:calibmethod}

We address the camera-to-mocap calibration problem: given a camera rigidly mounted on a mocap-tracked platform and a calibration board that is simultaneously visible to the camera and tracked by the mocap system, estimate the rigid transform from the platform's mocap frame to the camera frame. This is an instance of the $AX = YB$ hand-eye calibration problem~\cite{zhuang1994simultaneous}. Unlike prior work~\cite{sturm2012benchmark} that assumes the board-to-marker transform is precisely known, our formulation treats it as a free variable estimated jointly with the camera extrinsics.

\subsection{Problem Formulation}
\label{sec:calib_formulation}

Given a camera and a mocap system both observing a planar calibration board with attached retroreflective markers (\cref{fig:calboard}), the board is simultaneously detected in the camera's images and tracked in 3D by the mocap system.

\begin{figure}[t]
  \centering
  \includegraphics[width=0.95\linewidth]{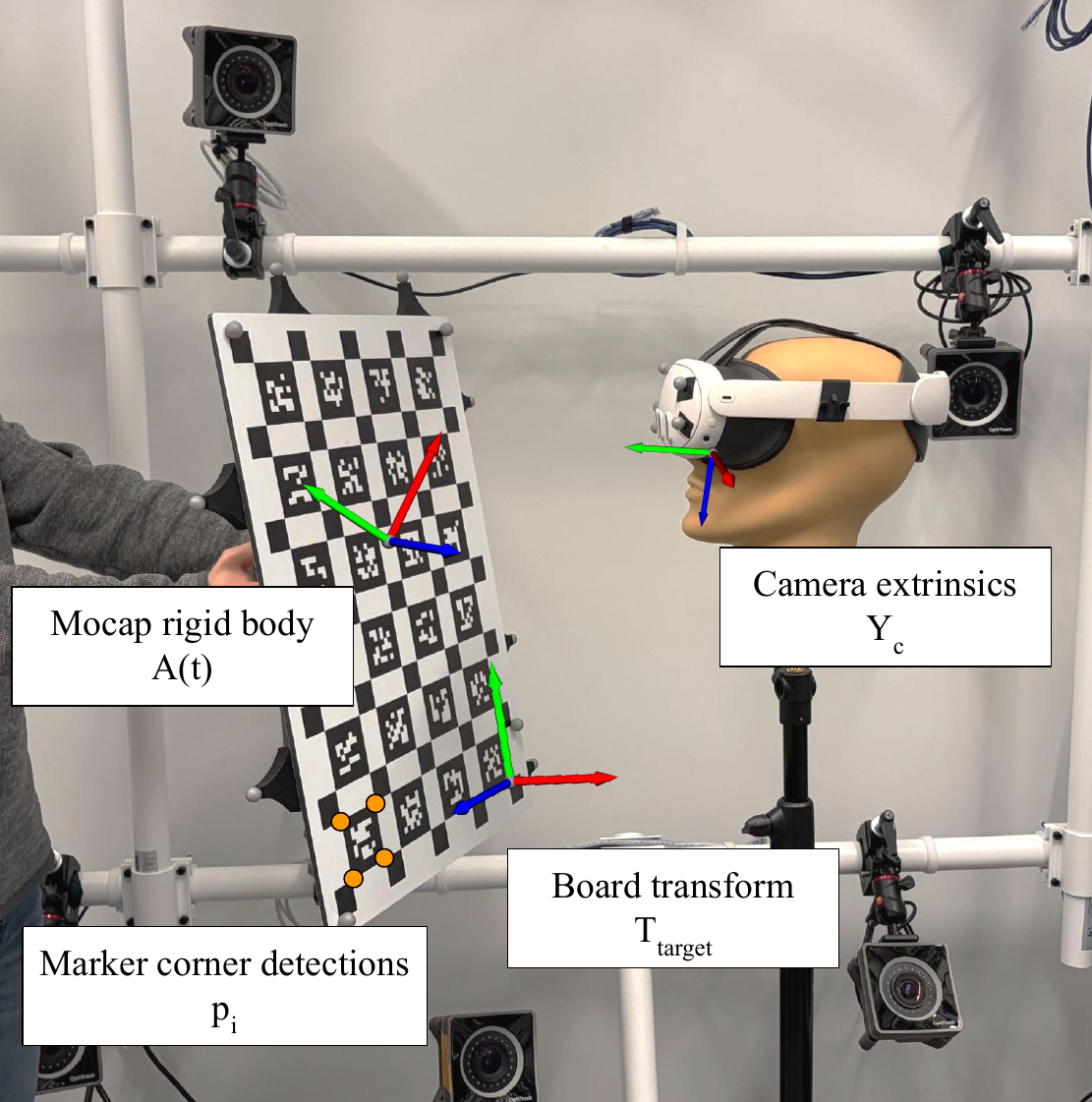}

    \caption{\textbf{Coordinate systems.} A mocap system tracks a rigid body mounted to the calibration board, $A(t)$. An ArUco solver estimates corner positions $\mathbf{p}_i$ and the board transform $T_{\mathrm{target}}$. Our calibration procedure solves for the headset extrinsics $Y_c$ and the transform between the mocap rigid body and the board transform, $X$.}
  
  \label{fig:coordinates}
\end{figure}

A 3D point $\mathbf{p}^\text{local}_i$ on the calibration target is projected into camera $c$ at frame $t$ via the following chain:
\begin{equation}
    \mathbf{u}_{c,t,i} = \pi\!\left(
    \underbrace{Y_c}_{\text{extrinsic}}
    \cdot\;
    \underbrace{A(t)}_{\text{mocap pose}}
    \cdot\;
    \underbrace{X}_{\text{board-to-marker}}
    \cdot\;
    \mathbf{p}^\text{local}_i
    ;\; \boldsymbol{\theta}_c\right)
    \label{eq:kinematic_chain}
\end{equation}
where:
\begin{itemize}\setlength{\itemsep}{3pt}
    \item $\mathbf{p}^\text{local}_i$ is the $i$-th corner coordinate in the target's local frame,
    \item $X = T^{\mathrm{rb}}_{\mathrm{target}}$ is the rigid transform (3 rotation + 3 translation DOFs) from the target frame to the mocap rigid body frame, shared across all cameras and frames,
    \item $A(t) = T^{\mathrm{world}}_{\mathrm{rb}}$ is the mocap-reported rigid body pose at frame $t$ (known),
    \item $Y_c = T^{\mathrm{eye}}_{\mathrm{world}}$ is the camera extrinsic, mapping mocap world coordinates into the camera frame,
    \item $\pi(\cdot; \boldsymbol{\theta}_c)$ is the camera projection function to produce pixel coordinates using intrinsic parameters $\boldsymbol{\theta}_c$.
\end{itemize}

Our goal is to estimate the unknown transforms $X$ and $\{Y_c\}$ for all cameras $c$, and optionally to refine the intrinsic parameters $\{\boldsymbol{\theta}_c\}$. In many settings, initial intrinsic estimates are already available from factory or online calibration (as on the Meta Quest~3) or from a separate intrinsic calibration step, and can be held fixed or used as a strong initialization.

\subsection{Optimization Objective}
\label{sec:calib_objective}

Classical approaches first estimate per-frame board poses via PnP, then solve $AX = YB$ from pairwise relative motions, discarding the original pixel observations. While our initialization also uses PnP to bootstrap the rigid transforms (\cref{sec:calib_solver}), the final objective is direct 2D reprojection error against the raw marker detections, cast as a bundle adjustment~\cite{triggs1999bundle,lourakis2009sba}. The residual for marker $i$ at frame $t$ observed by camera $c$ is:
\begin{equation}
    r_{c,t,i} =
    \pi\!\left(Y_c \cdot A(t) \cdot X \cdot \mathbf{p}^\text{local}_i;\; \boldsymbol{\theta}_c\right)
    - \mathbf{u}_{c,t,i}
    \label{eq:calib_residual}
\end{equation}
where $\mathbf{u}_{c,t,i}$ is the detected 2D corner. The full objective is:
\begin{equation}
    \min_{\{Y_c\},\, X,\, \{\boldsymbol{\theta}_c\}} \;\sum_{c,t,i} \left\| r_{c,t,i} \right\|^2
    \;+\; \sum_c \lambda \left\| \boldsymbol{\theta}_c - \boldsymbol{\theta}^*_c \right\|^2_W
    \label{eq:calib_objective}
\end{equation}
The regularization term penalizes deviation of intrinsic parameters from priors $\boldsymbol{\theta}^*_c$: the principal point is regularized toward the image center, distortion coefficients toward zero, and focal length is left effectively unregularized. This prevents degenerate solutions when detections are sparse.

\subsection{Initialization and Solution}
\label{sec:calib_solver}

Direct minimization of the 2D objective (\cref{eq:calib_objective}) is susceptible to local minima. We address this with a three-stage pipeline in which the first two stages operate on an auxiliary 3D point-matching objective that remains well-defined even for grossly incorrect estimates, and only the final stage optimizes the full 2D reprojection error.

\paragraph{3D point-matching objective.}
For each frame $t$ and camera $c$, we run PnP~\cite{lepetit2009epnp} on the 2D detections using the known marker geometry and initial intrinsic estimates $\boldsymbol{\theta}^*_c$ to recover a per-frame board pose in camera space. This yields reference 3D positions $\hat{\mathbf{p}}^\text{eye}_{c,t,i}$ for each detected marker $i$, computed once and held fixed throughout initialization. The 3D error measures how well the kinematic chain reproduces these references:
\begin{equation}
    E_\text{3D} = \sum_{c,t,i}
    \left\|
    Y_c \cdot A(t) \cdot X \cdot \mathbf{p}^\text{local}_i
    - \hat{\mathbf{p}}^\text{eye}_{c,t,i}
    \right\|^2
    \label{eq:3d_error}
\end{equation}
Unlike the 2D reprojection error, which requires points to lie in front of the camera and project within the image, $E_\text{3D}$ is well-defined for any configuration of the rigid transforms---even when the camera is pointed away from the board. This makes it suitable for coarse initialization where the current estimate may be far from the solution. The PnP-derived references do not participate in the final 2D objective; they serve only to guide the first two stages toward the correct basin of convergence. Intrinsic parameters are held fixed during both stages.

\paragraph{Stage 1: Random-restart Procrustes.}
The board-to-marker transform $X$ is the most critical variable to initialize correctly. Because the orientation of the mocap rigid body frame relative to the printed pattern is unknown \emph{a priori}, $X$ may be initialized with an arbitrarily wrong rotation. If, for example, $X$ is off by 180\textdegree, each camera independently concludes that the markers it observes lie on the reverse side of the board and places itself on the wrong side of the room---a deep local minimum from which gradient-based solvers cannot escape.

To resolve this ambiguity, we sample 30 well-distributed candidate rotations for $X$ via a greedy farthest-point algorithm on $30 \times 10 = 300$ uniform random rotations, using Frobenius distance $d(R_i, R_j) = \|R_i - R_j\|_F$. For each candidate, we initialize the rotation of $X$ and run iterative Procrustes alignment: each transform in the chain is solved in turn via closed-form SVD-based rigid registration~\cite{arun1987least} on the 3D correspondences from \cref{eq:3d_error}, with the board-to-marker transform $X$ solved last to prevent it from absorbing global rigid body motion. We select the candidate with the lowest $E_\text{3D}$. This stage requires no iterative matrix inversion and reliably identifies the correct rotation basin.

\paragraph{Stage 2: 3D error minimization.}
Stage~1 solves each transform independently in closed form, which is fast but does not account for interactions between transforms. Using the best Procrustes initialization, a Gauss-Newton solver with Levenberg-Marquardt damping~\cite{marquardt1963algorithm} minimizes $E_\text{3D}$ (\cref{eq:3d_error}), jointly refining all rigid transforms simultaneously. This yields a geometrically accurate estimate suitable for initializing the 2D reprojection stage.

\paragraph{Stage 3: 2D reprojection refinement.}
Using the Stage~2 result as initialization, the same solver minimizes the full 2D objective (\cref{eq:calib_objective}), jointly optimizing rigid transforms and camera intrinsics against pixel-level detections. Iteration terminates when improvement drops below $\varepsilon(1 + E)$ with $\varepsilon = 10^{-4}$.

The three-stage approach ensures global convergence (Stage~1), geometric accuracy (Stage~2), and pixel-level precision (Stage~3). Rotations are parameterized via the exponential map~\cite{grassia1998practical} with 6-DOF (3 rotation + 3 translation) incremental updates; all Jacobians are computed analytically.

\section{Verification Method}
\label{sec:verification}

\begin{figure}[t]
  \centering
  \includegraphics[width=0.98\linewidth]{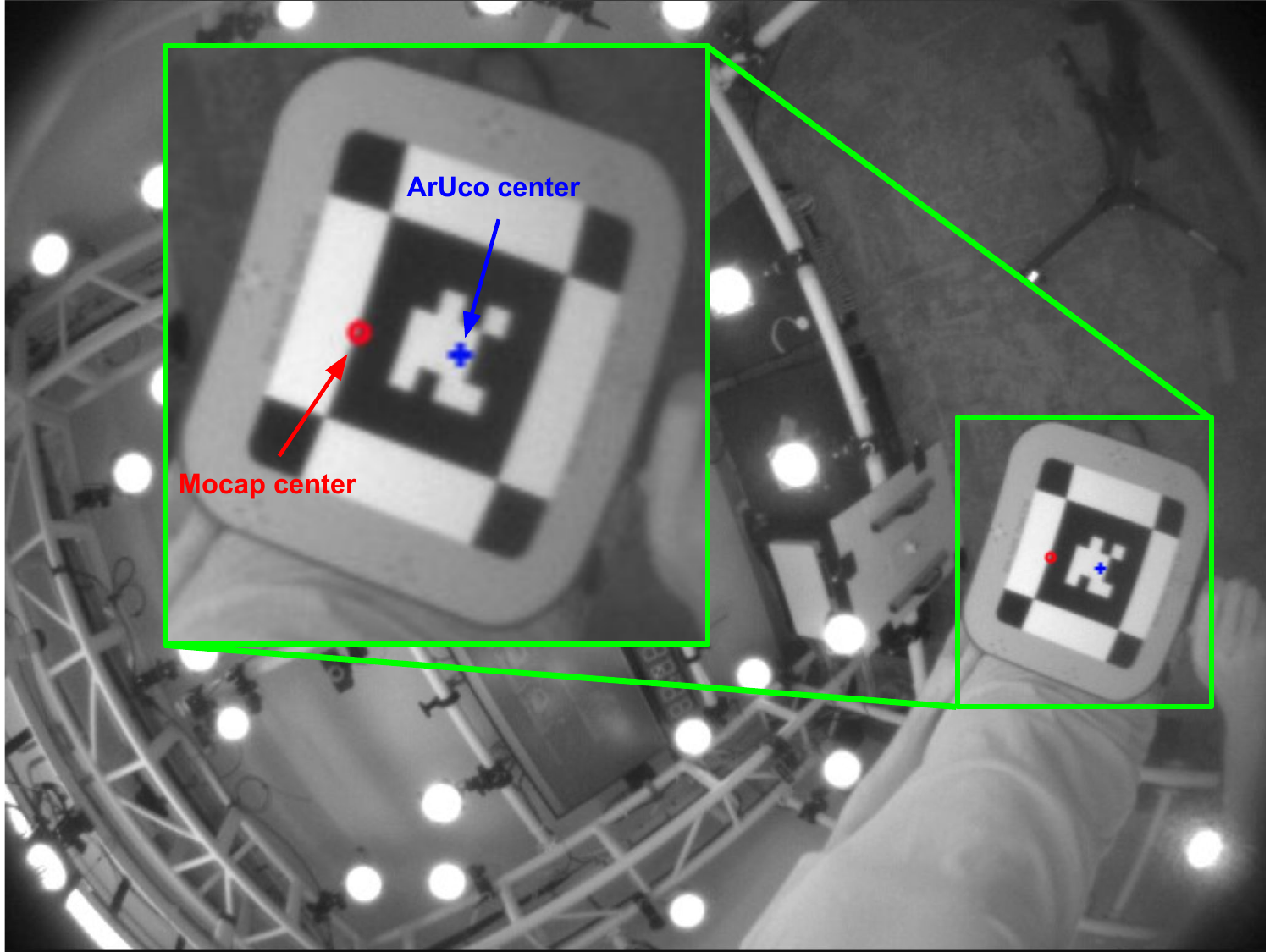}
  \caption{\textbf{\lollypop verification in action.} A fisheye camera image with the detected ArUco board center (blue crosshair) and the mocap rigid body centroid projected through the calibration (red circle). When calibration is correct, the two markers coincide.}
  \label{fig:aruco_vs_mocap}
\end{figure}

We present \lollypop, a device and method for independently verifying camera-to-mocap calibration. The key idea is to compare two independent estimates of a device's image position---one from fiducial detection, one from mocap projection through the calibration---using a separate recording collected after calibration is complete.

\subsection{Physical Device}
\label{sec:device}

The \lollypop device consists of two rigidly coupled components: an ArUco~\cite{garrido2014automatic,romero2018speeded} fiducial marker board and a mocap rigid body comprising six retroreflective markers. The board is printed on adhesive-backed polypropylene and mounted to a rigid backing panel. The mocap markers are 8\,mm circular sections of 3M Scotchlite 7610 retroreflective tape applied directly to the board surface. The ArUco board provides the image-space reference, while the mocap rigid body provides the 3D tracking-space reference.

The rigid body uses six mocap markers rather than the minimum three for two reasons. First, the centroid of $N$ markers has tracking noise reduced by a factor of $\sqrt{N}$ relative to a single marker, improving 3D position precision.

Critically, the device is constructed so that the ArUco board center and the mocap rigid body centroid are \emph{coincident} by design, i.e., the 3D offset is 0. This is a fixed mechanical constraint of the rigid device. Unlike the calibration board's board-to-marker attachment---which can shift between sessions and is estimated as part of the calibration solve (\cref{sec:calibmethod})---the Lollypop's geometric relationship does not drift and is not re-estimated. This means projecting the mocap rigid body centroid through the calibration produces exactly the predicted image position of the ArUco board center.


\subsection{Device Precision}
\label{sec:precision}




The coincidence of the ArUco board center and mocap rigid body centroid is verified by direct measurement against the CAD design. The primary source of manufacturing error is the thickness of the retroreflective tape (0.12\,mm), which raises the mocap marker centroids above the printed board plane. This height difference contributes zero error when the board is viewed at normal incidence (zenith) and a maximum of 0.12\,mm at 90\textdegree{} grazing angle. In practice, ArUco detection is inherently limited to moderate viewing angles, bounding this error well below 0.12\,mm; additional filtering on the detection viewing angle can further reduce its impact for tighter tolerance requirements.


\subsection{Error Metric}
\label{sec:metric}

For each frame where both the ArUco marker and the mocap rigid body are detected, we compare two independent estimates of the device center in the image (\cref{fig:aruco_vs_mocap}) in both 2D and 3D.

\paragraph{ArUco center.}
The board center is estimated from the four detected corners via a 2D homography. Each corner is undistorted to normalized image coordinates using the camera model.
A homography $\vect{H}$ from a canonical unit square to these normalized positions is estimated, and the canonical center $(0,0)$ is warped through $\vect{H}$ and mapped back to distorted pixel coordinates to yield $\vect{p}_\text{aruco}$. This operates entirely in the image domain without 3D pose estimation.

\paragraph{Mocap center.}
Because the board center and rigid body centroid are coincident (\cref{sec:device}), the mocap centroid $\vect{X}_\text{mocap}$ projects directly as the predicted board center:
\begin{equation}
    \vect{p}_\text{mocap} = \pi\!\left(\vect{R} \cdot \vect{X}_\text{mocap} + \vect{t};\; \boldsymbol{\theta}_c\right)
    \label{eq:mocap_proj}
\end{equation}
where $[\vect{R} | \vect{t}]$ is the world-to-camera extrinsic, composed with the inverse platform pose for headset-mounted cameras.

\paragraph{2D error.}
The primary metric is the pixel-space Euclidean distance:
\begin{equation}
    e_\text{2D} = \norm{\vect{p}_\text{aruco} - \vect{p}_\text{mocap}}_2
    \label{eq:error_2d}
\end{equation}
Both points are in distorted pixel coordinates, so this tests the full calibration chain including the distortion model.

\paragraph{3D error.}
We also back-project $\vect{p}_\text{aruco}$ to 3D at the depth of the projected mocap centroid, yielding a world-space ArUco position $\vect{X}_\text{aruco}$:
\begin{equation}
    e_\text{3D} = \norm{\vect{X}_\text{mocap} - \vect{X}_\text{aruco}}_2
    \label{eq:error_3d}
\end{equation}
This gives error in millimeters but is not fully independent in the depth direction. We report both; 2D error is our primary measure.



\section{Experiments}
\label{sec:experiments}

We evaluate on a Meta Quest~3 headset equipped with four fisheye cameras using the Fisheye62 distortion model, which extends Kannala-Brandt~\cite{kannala2006generic} with two tangential distortion coefficients (6 radial + 2 tangential parameters). Camera intrinsics are read directly from the device's online calibration and held fixed; the solver optimizes only extrinsics and the board-to-marker transform. Headset camera images are hardware-synchronized with the mocap system, which provides rigid body poses of the calibration board and headset.
A printed A2 fiducial board with 24 ArUco markers is used as the detection board. Each marker has a side length of 59 mm, with 29.5 mm spacing between neighboring markers. The layout also includes smaller black square patches at the grid intersections (and boundary intersections), which strengthens local contrast and helps stabilize corner localization. The four corners of each marker are used as known 3D reference points in the board coordinate frame during calibration. Our custom board follows the ChArUco principle of combining ArUco markers with checkerboard-style corner structure for robust corner localization~\cite{garrido2014automatic,romero2018speeded}.

\subsection{Calibration Benchmark}
\label{sec:calib_benchmark}

We compare our joint calibration (\cref{sec:calibmethod}) against Sturm~\etal~\cite{sturm2012benchmark} on five calibration board recordings. Both methods receive identical input: detected board corners and mocap rigid body poses of the board and headset. Sturm~\etal requires the board-to-marker transform $X$ to be provided as a known input; our method estimates it jointly. We report both the self-referential board RMSE (reprojection error on calibration data) and the independent \lollypop 2D error from a separate verification recording.

Since no open-source implementation of Sturm~\etal is available, we implement an approximation of their method.
Using the calibration board shown in \cref{fig:calboard}, we place four retroreflective markers at the corners of ArUco tags, forming a rigid body whose centroid-to-board-center offset is measured and held fixed as the known board-to-marker transform~$X$.
Given this fixed~$X$, the reimplemented baseline solves for the camera extrinsic by minimizing reprojection error through the kinematic chain and the camera intrinsic model.
The solver proceeds in two stages: an initial least-squares fit for basin capture, followed by robust refinement with Huber-weighted residuals~\cite{huber1964robust}.
In contrast, our method uses additional markers affixed to the outer perimeter of the board (\cref{fig:calboard}) without requiring precise placement;
the board-to-marker transform~$X$ is treated as a free variable and estimated jointly with the camera extrinsics (\cref{sec:calibmethod}), eliminating the need for error-prone manual measurement.

We collect five calibration recordings with the headset placed at different poses within the capture volume to vary the viewing geometry.
For each recording, both methods are evaluated on the same set of detection results and mocap poses.
We report the board reprojection RMSE, computed as the pooled per-point root-mean-square error over all detected corners:
\begin{equation}
  \text{RMSE} = \sqrt{\frac{1}{N}\sum_{i=1}^{N}\left\|r_i\right\|^2}
  \label{eq:board_rmse}
\end{equation}
where $r_i$ is the 2D reprojection residual (\cref{eq:calib_residual}) and $N$ is the total number of detected corner observations across all frames.

\begin{table}[t]
\centering
\caption{\textbf{Calibration benchmark: Ours vs.\ Sturm~\etal~\cite{sturm2012benchmark}.} Board RMSE is the self-referential calibration residual; \lollypop RMSE is the independent verification metric. Per camera, aggregated across 5 recordings. Lower is better.}
\label{tab:calib_benchmark}
\small
\resizebox{\columnwidth}{!}{%
\begin{tabular}{@{}lcccc@{}}
\toprule
 & \multicolumn{2}{c}{Board RMSE (px)} & \multicolumn{2}{c}{\lollypop RMSE (px)} \\
\cmidrule(lr){2-3} \cmidrule(lr){4-5}
Camera & Ours & Sturm & Ours & Sturm \\
\midrule
Right Front  & \textbf{0.12} & 1.04 & \textbf{0.22} & 1.87 \\
Left Front   & \textbf{0.12} & 0.40 & \textbf{0.24} & 1.91 \\
Left Side    & \textbf{0.08} & 0.81 & \textbf{0.44} & 1.76 \\
Right Side   & \textbf{0.09} & 6.76 & \textbf{0.35} & 8.45 \\
\midrule
\textbf{Mean} & \textbf{0.10} & 2.25 & \textbf{0.31} & 3.50 \\
\bottomrule
\end{tabular}}
\end{table}

\cref{tab:calib_benchmark} shows that our method produces lower \lollypop error compared to the approximated Sturm~\etal method across all five recordings.

\subsection{Ablation Study}
\label{sec:ablation}

We ablate key components of the optimization pipeline to measure their individual contributions.
Starting from the full method, we disable each feature in isolation and evaluate the resulting board reprojection RMSE on a representative calibration recording.

\begin{table}[ht]
\centering
\caption{\textbf{Ablation study.} Effect of disabling individual optimization components on board reprojection RMSE (px). Each row removes one component from the full pipeline.}
\label{tab:ablation}
\small
\resizebox{\columnwidth}{!}{%
\begin{tabular}{@{}lcc@{}}
\toprule
Configuration & RMSE (px) & $\Delta$ \\
\midrule
Full pipeline (Stages 1+2+3)        & 0.146  & --- \\
w/o Stage 3 (2D reprojection)       & 0.148  & +0.002 \\
w/o Procrustes initialization       & 0.146  & +0.000 \\
Stage 1 only (no iterative refine.) & 43.000 & +42.900 \\
\bottomrule
\end{tabular}}
\end{table}

\cref{tab:ablation} reveals that each stage serves a distinct role.
Without any iterative refinement (Stage~1 only), error is ${\sim}300\times$ worse (43\,px), confirming that iterative optimization is essential.
\textbf{Procrustes initialization} (Stage~1) does not improve final accuracy but resolves the rotation ambiguity in~$X$, reducing Stage~2 iterations from 11 to~6 ($1.8\times$ convergence speedup).
Its primary contribution is solver robustness: ensuring the Gauss-Newton solver lands in the correct basin,
which is critical when the board-to-marker transform is far from identity or the board has near-symmetry.
\textbf{Stage~2} (3D error minimization) provides well-conditioned convergence from the coarse Procrustes initialization.
Its smoother objective landscape, free of projection nonlinearity, reliably drives large initial errors down to the geometric minimum.
\textbf{Stage~3} (2D reprojection refinement) jointly refines intrinsics and achieves optimal pixel-level fit.
On this recording the improvement over Stage~2 alone is marginal (0.146 vs.\ 0.148\,px) because we fixed the intrinsics for controlled experiments;
on recordings with poor factory calibration, Stage~3 is expected to contribute more substantially, reflecting the intrinsics adjustment that improves pixel fit while shifting 3D reconstruction.

\begin{figure*}[!t]
  \centering
  \begin{subfigure}{0.48\linewidth}
    \includegraphics[width=\linewidth]{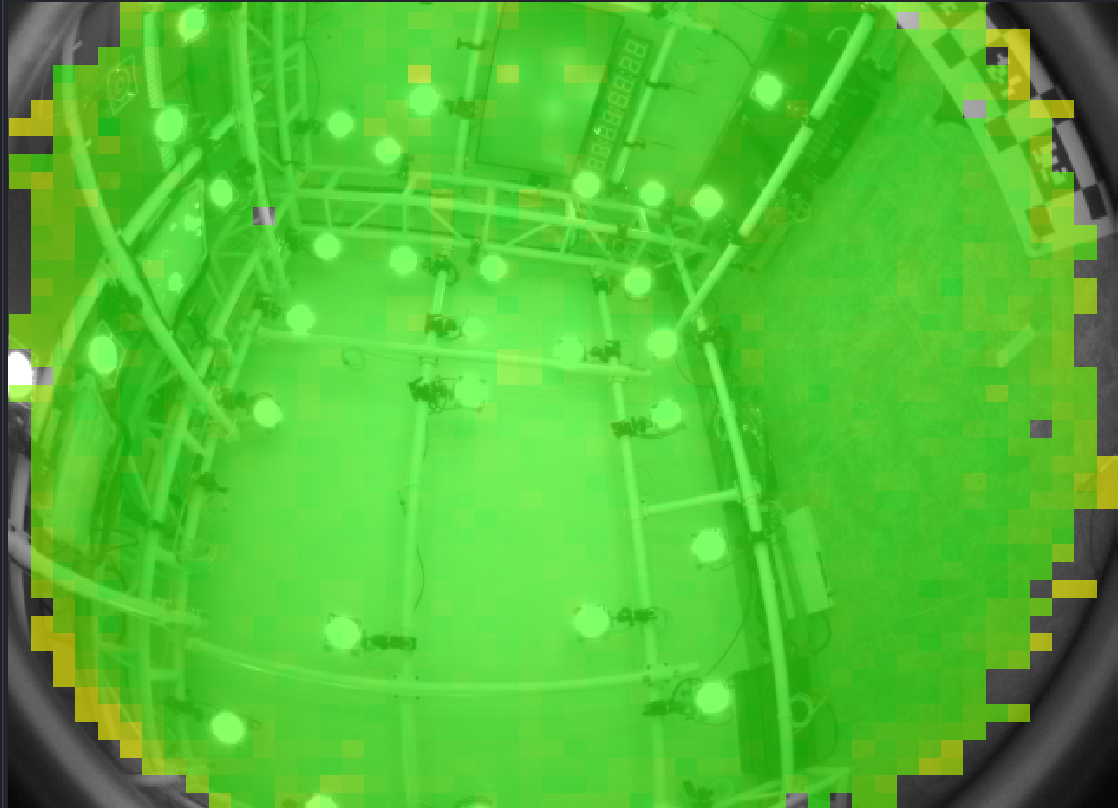}
    \caption{High-quality calibration}
    \label{fig:heatmap-good}
  \end{subfigure}
  \hfill
  \begin{subfigure}{0.48\linewidth}
    \includegraphics[width=\linewidth]{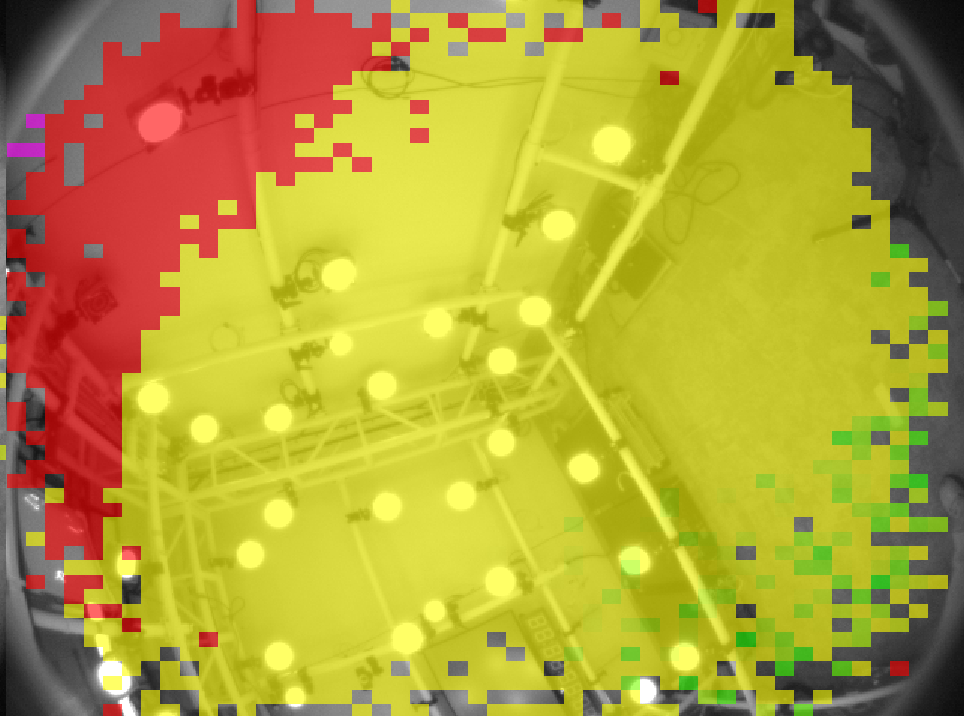}
    \caption{Low-quality calibration}
    \label{fig:heatmap-bad}
  \end{subfigure}
  \caption{\textbf{Spatial error heatmaps: high-quality vs.\ low-quality calibration on the same fisheye camera.} \lollypop 2D reprojection error binned by image position and averaged within each bin. Color encodes per-bin mean error: green ($<0.5$\,px), yellow ($0.5$--$1.5$\,px), red ($1.5$--$3.0$\,px), magenta ($>3.0$\,px). The high-quality calibration~(a) shows uniformly low error (green). The low-quality calibration~(b) shows error at non-uniform locations.}
  \label{fig:heatmaps}
\end{figure*}

\subsection{Calibration Drift Over Time}
\label{sec:drift}

A key motivation for independent verification is detecting calibration degradation that accumulates during normal use. \cref{fig:drift} shows error measured over five successive verification recordings. Between recordings, the headset is repeatedly donned and doffed, and retroreflective markers are occasionally bumped or shifted to emulate real-world handling. Both pixel-domain and metric-domain RMSE increase over repeated use, demonstrating that a calibration performed initially can degrade overtime. \lollypop provides a per-session quick check that detects when recalibration is needed. Visualization of the mocap vs ArUco center on the raw image as shown in \cref{fig:aruco_vs_mocap} also provides a way to spot check for any offset.

\subsection{Spatially-Resolved Error Analysis}
\label{sec:spatial}

Aggregate metrics report a single number per camera, which is especially misleading for fisheye cameras where calibration quality varies across the image. \cref{fig:heatmaps} shows \lollypop error binned by image position for the same camera under accurate and inaccurate calibrations. The high-quality calibration exhibits uniformly low error across the image. The bad calibration shows lower error near the optical center and elevated error at the top/left periphery.

\section{Conclusion}
\label{sec:conclusion}

We have presented a complete calibrate-then-verify pipeline for camera-to-mocap systems with fisheye cameras. Our calibration method formulates the problem as a kinematic-chain bundle adjustment that jointly solves for camera extrinsics and the board-to-marker transform, eliminating the assumption that the latter is precisely known. Together with staged initialization and refinement, this improves robustness to attachment variation and initialization ambiguity. Benchmarked against Sturm~\etal~\cite{sturm2012benchmark}, our method produces substantially lower independent verification error across all cameras.

For verification, \lollypop provides an independent calibration check using a physical device that combines fiducial markers with mocap rigid bodies. In controlled experiments on fisheye cameras, \lollypop reliably offered observability into calibration degradation. Spatially-resolved error analysis reveals non-uniform calibration quality that aggregate metrics mask entirely: errors concentrate at the image periphery where lens distortion is strongest. Together with the repeated-use drift analysis, this supports robustness not only at calibration time, but also for deployment-time quality control.

The current approach requires optical mocap infrastructure for the 3D reference, and the 2D error metric primarily validates lateral calibration accuracy---depth-direction errors are less well constrained since the back-projection uses mocap-derived depth. A future extension using 3D fiducial arrangements (\eg, markers on multiple faces of a polyhedron) could address both limitations by improving viewing-angle coverage and enabling depth-direction verification.

In our production data collections, \lollypop cut validation time by over 50\% and replaced subjective visual inspection with an objective pass/fail criterion. More broadly, we hope this paper helps shift calibration from an under-documented preprocessing step to a rigorously measured part of large-scale dataset construction, improving the quality and reproducibility of future datasets.

{
    \small
    \bibliographystyle{ieeenat_fullname}
    \bibliography{main}

@String(PAMI  = {IEEE TPAMI})

@String(IJCV  = {IJCV})

@String(CVPR  = {CVPR})

@String(ICCV  = {ICCV})

@String(ECCV  = {ECCV})

@String(IROS  = {IEEE/RSJ IROS})

@article{zhang2000flexible,
  title={A flexible new technique for camera calibration},
  author={Zhang, Zhengyou},
  journal=PAMI,
  volume={22},
  number={11},
  pages={1330--1334},
  year={2000}
}

@article{tsai1987versatile,
  title={A versatile camera calibration technique for high-accuracy {3D} machine vision metrology using off-the-shelf {TV} cameras and lenses},
  author={Tsai, Roger},
  journal={IEEE J. Robotics and Automation},
  volume={3},
  number={4},
  pages={323--344},
  year={1987}
}

@inproceedings{scaramuzza2006toolbox,
  title={A toolbox for easily calibrating omnidirectional cameras},
  author={Scaramuzza, Davide and Martinelli, Agostino and Siegwart, Roland},
  booktitle=IROS,
  pages={5695--5701},
  year={2006}
}

@inproceedings{tezaur2022new,
  title={A new non-central model for fisheye calibration},
  author={Tezaur, Radka and Kumar, Avinash and Nestares, Oscar},
  booktitle=CVPR,
  pages={5222--5231},
  year={2022}
}

@article{kannala2006generic,
  title={A generic camera model and calibration method for conventional, wide-angle, and fish-eye lenses},
  author={Kannala, Juho and Brandt, Sami S},
  journal=PAMI,
  volume={28},
  number={8},
  pages={1335--1340},
  year={2006}
}

@inproceedings{furgale2013unified,
  title={Unified temporal and spatial calibration for multi-sensor systems},
  author={Furgale, Paul and Rehder, Joern and Siegwart, Roland},
  booktitle=IROS,
  pages={1280--1286},
  year={2013}
}

@article{lepetit2009epnp,
  title={{EPnP}: An accurate {O(n)} solution to the {PnP} problem},
  author={Lepetit, Vincent and Moreno-Noguer, Francesc and Fua, Pascal},
  journal=IJCV,
  volume={81},
  number={2},
  pages={155--166},
  year={2009}
}

@article{arun1987least,
  title={Least-squares fitting of two {3-D} point sets},
  author={Arun, K Somani and Huang, Thomas S and Blostein, Steven D},
  journal=PAMI,
  number={5},
  pages={698--700},
  year={1987}
}

@article{marquardt1963algorithm,
  title={An algorithm for least-squares estimation of nonlinear parameters},
  author={Marquardt, Donald W},
  journal={J. Society for Industrial and Applied Mathematics},
  volume={11},
  number={2},
  pages={431--441},
  year={1963}
}

@inproceedings{triggs1999bundle,
  title={Bundle adjustment---a modern synthesis},
  author={Triggs, Bill and McLauchlan, Philip F and Hartley, Richard I and Fitzgibbon, Andrew W},
  booktitle={Int. Workshop on Vision Algorithms},
  pages={298--372},
  year={1999}
}

@article{lourakis2009sba,
  title={{SBA}: A software package for generic sparse bundle adjustment},
  author={Lourakis, Manolis I A and Argyros, Antonis A},
  journal={ACM Trans. Mathematical Software},
  volume={36},
  number={1},
  pages={1--30},
  year={2009}
}

@article{grassia1998practical,
  title={Practical parameterization of rotations using the exponential map},
  author={Grassia, F Sebastian},
  journal={J. Graphics Tools},
  volume={3},
  number={3},
  pages={29--48},
  year={1998}
}

@article{huber1964robust,
  title={Robust estimation of a location parameter},
  author={Huber, Peter J},
  journal={Annals of Mathematical Statistics},
  volume={35},
  number={1},
  pages={73--101},
  year={1964}
}

@inproceedings{schofield2018calibration,
  title={Calibration for camera-motion capture extrinsics},
  author={Schofield, Sam D and Edwards, Matthew J and Green, Richard D},
  booktitle={Int. Conf. Image and Vision Computing New Zealand (IVCNZ)},
  pages={1--6},
  year={2018}
}

@article{shu2024spatiotemporal,
  title={A spatiotemporal hand-eye calibration for trajectory alignment in visual(-inertial) odometry evaluation},
  author={Shu, Zhan and Bei, Siyu and Dai, Jinhao and Li, Lin and Chen, Zheng and Zhang, Hui},
  journal={IEEE Robotics and Automation Letters},
  volume={9},
  number={6},
  pages={5134--5141},
  year={2024}
}

@inproceedings{richardson2013aprilcal,
  title={{AprilCal}: Assisted and repeatable camera calibration},
  author={Richardson, Andrew and Strom, Johannes and Olson, Edwin},
  booktitle=IROS,
  pages={4618--4624},
  year={2013}
}

@article{zhuang1994simultaneous,
  title={Simultaneous robot/world and tool/flange calibration by solving homogeneous transformation equations of the form {$AX = YB$}},
  author={Zhuang, Hanqi and Roth, Zvi S and Sudhakar, Raghavan},
  journal={IEEE Trans. Robotics and Automation},
  volume={10},
  number={4},
  pages={549--554},
  year={1994}
}

@article{dornaika1998simultaneous,
  title={Simultaneous robot-world and hand-eye calibration},
  author={Dornaika, Fadi and Horaud, Radu},
  journal={IEEE Trans. Robotics and Automation},
  volume={14},
  number={4},
  pages={617--622},
  year={1998}
}

@inproceedings{chiodini2018camera,
  title={Camera rig extrinsic calibration using a motion capture system},
  author={Chiodini, Sebastiano and Pertile, Marco and Giubilato, Riccardo and Salvioli, Federico and Barrera, Marco and Franceschetti, Paola and Debei, Stefano},
  booktitle={IEEE Int. Workshop on Metrology for AeroSpace},
  pages={590--595},
  year={2018}
}

@inproceedings{sturm2012benchmark,
  title={A benchmark for the evaluation of {RGB-D} {SLAM} systems},
  author={Sturm, J{\"u}rgen and Engelhard, Nikolas and Endres, Felix and Burgard, Wolfram and Cremers, Daniel},
  booktitle=IROS,
  pages={573--580},
  year={2012}
}

@article{tsai1989new,
  title={A new technique for fully autonomous and efficient {3D} robotics hand/eye calibration},
  author={Tsai, Roger Y and Lenz, Reimar K},
  journal={IEEE Trans. Robotics and Automation},
  volume={5},
  number={3},
  pages={345--358},
  year={1989}
}

@article{garrido2014automatic,
  title={Automatic generation and detection of highly reliable fiducial markers under occlusion},
  author={Garrido-Jurado, Sergio and Mu{\~n}oz-Salinas, Rafael and Madrid-Cuevas, Francisco Jos{\'e} and Mar{\'\i}n-Jim{\'e}nez, Manuel Jes{\'u}s},
  journal={Pattern Recognition},
  volume={47},
  number={6},
  pages={2280--2292},
  year={2014}
}

@article{romero2018speeded,
  title={Speeded up detection of squared fiducial markers},
  author={Romero-Ramirez, Francisco J and Mu{\~n}oz-Salinas, Rafael and Medina-Carnicer, Rafael},
  journal={Image and Vision Computing},
  volume={76},
  pages={38--47},
  year={2018}
}

@inproceedings{schubert2018tumvi,
  title={The {TUM VI} benchmark for evaluating visual-inertial odometry},
  author={Schubert, David and Goll, Thore and Demmel, Nikolaus and Usenko, Vladyslav and St{\"u}ckler, J{\"o}rg and Cremers, Daniel},
  booktitle=IROS,
  pages={1680--1687},
  year={2018}
}

@article{burri2016euroc,
  title={The {EuRoC} micro aerial vehicle datasets},
  author={Burri, Michael and Nikolic, Janosch and Gohl, Pascal and Schneider, Thomas and Rehder, Joern and Omari, Sammy and Achtelik, Markus W and Siegwart, Roland},
  journal={Int. J. Robotics Research},
  volume={35},
  number={10},
  pages={1157--1163},
  year={2016}
}

@article{ionescu2014human36m,
  title={Human3.6{M}: Large scale datasets and predictive methods for {3D} human sensing in natural environments},
  author={Ionescu, Catalin and Papava, Dragos and Olaru, Vlad and Sminchisescu, Cristian},
  journal=PAMI,
  volume={36},
  number={7},
  pages={1325--1339},
  year={2014}
}

@inproceedings{sener2022assembly101,
  title={{Assembly101}: A large-scale multi-view video dataset for understanding procedural activities},
  author={Sener, Fadime and Chatterjee, Dibyadip and Shelepov, Daniel and He, Kun and Singhania, Dipika and Wang, Robert and Yao, Angela},
  booktitle=CVPR,
  pages={21064--21074},
  year={2022}
}

@inproceedings{zimmermann2019freihand,
  title={{FreiHAND}: A dataset for markerless capture of hand pose and shape from single {RGB} images},
  author={Zimmermann, Christian and Ceylan, Duygu and Yang, Jimei and Russell, Bryan and Argus, Max and Brox, Thomas},
  booktitle=ICCV,
  pages={813--822},
  year={2019}
}

@inproceedings{moon2020interhand,
  title={{InterHand2.6M}: A dataset and baseline for {3D} interacting hand pose estimation from a single {RGB} image},
  author={Moon, Gyeongsik and Yu, Shoou-I and Wen, He and Shiratori, Takaaki and Lee, Kyoung Mu},
  booktitle=ECCV,
  pages={548--564},
  year={2020}
}

@inproceedings{fan2023arctic,
  title={{ARCTIC}: A dataset for dexterous bimanual hand-object manipulation},
  author={Fan, Zicong and Taheri, Omid and Tzionas, Dimitrios and Kocabas, Muhammed and Kaufmann, Manuel and Black, Michael J and Hilliges, Otmar},
  booktitle=CVPR,
  pages={12943--12954},
  year={2023}
}
}

\end{document}